\documentclass[runningheads]{llncs}
\pdfoutput=1

\usepackage[utf8]{inputenc} 
\usepackage[english]{babel}
\usepackage{amsmath}
\usepackage{graphicx} 
\usepackage{xspace} 
\usepackage{todonotes}
\usepackage{subfig} 
\usepackage{pdflscape}

\newcommand{\contentPath}{./}
\newcommand{\imagePath}{./}

\usepackage{amssymb}

\makeatletter
\newcommand{\plusminus}{\mathbin{\mathpalette\@plusminus\relax}}
\newcommand{\@plusminus}[2]{\ooalign{%
  \raisebox{.1\height}{$#1+$}\cr
  \smash{\raisebox{-.6\height}{$#1-$}}\cr}}
\makeatother

\usepackage[pdftex=true, citecolor=green, colorlinks=true, breaklinks=true, linkcolor=black, menucolor=black, urlcolor=black]{hyperref} 

\begin{document}
\title{Structural Analysis of Sparse Neural Networks}
%
%
\author{Julian Stier \and Michael Granitzer}
\authorrunning{J. Stier et al.}
\institute{University of Passau, Innstraße 41, 94032 Passau\protect\\\email{forename.surname@uni-passau.de}}

\maketitle

\begin{abstract}
Sparse Neural Networks regained attention due to their potential of mathematical and computational advantages.
We give motivation to study Artificial Neural Networks (ANNs) from a network science perspective, provide a technique to embed arbitrary Directed Acyclic Graphs into ANNs and report study results on predicting the performance of image classifiers based on the structural properties of the networks' underlying graph.
Results could further progress neuroevolution and add explanations for the success of distinct architectures from a structural perspective.
\end{abstract}

\section{Introduction}\label{sec:introduction}
Artificial Neural Networks (ANNs) are highly successful machine learning models and achieve human performance in various domains such as image processing and speech synthesis.
The choice on architecture is crucial for their success -- but ANN architectures, seen as network structures, have not been extensively studied from a network science perspective, yet.
The motivation of studying ANNs from a network science perspective is manifold:

First of all, ANNs are inspired from biological neural networks (notably the human brain) which have scale-free properties and ``are shown to be small-world networks'' \cite{watts1998collective}.
This is contradictory to most successful models in various machine learning problem domains.
However, with respect to their required neurons, those successful models have shown to be redundant by a magnitude.
For example, Han et al. claim ``on the ImageNet dataset, our method reduced the number of parameters of AlexNet by a factor of 9$\times$'' and ``VGG-16 can be reduced by 13$\times$'' \cite{han2015learning} through pruning.
Thus, biology and structural searches such as e.g. pruning suggest to develop sparse neural network structures \cite{stier2018analysing}.

Secondly, a look on the history of ANNs shows that, after major breakthroughs in the early 90s, the research community already tried to find characteristic networks structures by e.g. constructive and destructive approaches. 
However, network structures have only been studied in graph theory by then and the major remaining architectures from this research period are notably Convolutional Neural Networks (CNNs) \cite{lecun1990handwritten} and Long Short-Term Memory Networks (LSTMs) \cite{hochreiter1997long}.
New breakthroughs in the early 21st century led to recent trends of new architectures such as Highway Networks \cite{srivastava2015highway} and Residual Networks \cite{he2016deep}.
This suggests that further insights from revisiting and analysing the topology of sparse neural networks could be gained, particularly from a network science perspective as done in Mocanu et al. \cite{mocanu2017evolutionary}, which also ``argue that ANNs, too, should not have fully-connected layers''.

Last but not least, many researchers reported various advantages of sparse structures over unjustified densely stacked layers.
Glorot et al. argued, that sparsity is one of the factors for the success of Rectified Linear Units (ReLU) as activation functions because ``using a rectifying non-linearity gives rise to real zeros of activations and thus truly sparse representations'' \cite{glorot2011deep}.
The optimization process with ReLUs might not only find good representations but also better structures through sparser connections.
Sparser connections can also decrease the number of computations and thus time and energy consumption.
Not only exists biological motivation but also mathematical and computational advantages have been reported for sparse neural network structures.

Looking at the structural design and training of ANNs as a search problem, one can argue to apply \textbf{structural regularisation} when postulating distinct structural properties.
CNNs exploit sparsely connected neurons due to spatial relationships of the input features, which leads in conjunction with weight-sharing to very successful models.
LSTMs overcome analytical issues in training by postulating structures with slightly changed training behaviour.
We argue, that there can be new insights gained from studying ANNs from a network science perspective.

Sparse Neural Networks (SNNs), which we define as networks not being fully connecteted between layers, form another important, yet not well-understood structural regularisation.
We mention three major approaches to study the influence of structural regularisation to properties of SNNs: 1) studying fixed sparse structures with empirical success or selected analytical advantages, 2) studying sparse structures obtained in search heuristics and 3) studying sparse structures with common graph theoretical properties.

The first approach studies selected models with fixed structure analytically or within empirical works.
Exemplarily, LSTMs arose from analytical work \cite{hochreiter1991untersuchungen} on issues in the domain of time-dependent problems and have also been studied empirically since then.
In larger network structures, LSTMs and CNNs provide fixed components which have been proven to be successful due to analytical research and much experience in empirical work.
These insights provide foundations to construct larger networks in a bottom-up fashion.

On larger scale, sparse network structures can be approached by automatic construction, pruning, evolutionary techniques or even as a result of training regularisation or the choice of activation function.
Despite the common motivation to find explanations for resulting sparse structures, a lot of those approaches indeed obtain sparse structures but fail to find an explanation for them.

Studying real-world networks has been addressed by the field of Network Science.
However, as of now, Network Science has been hardly used to study the structural properties of ANNs and to understand the regularisation effects introduced by sparse network structures.

~\\
\noindent This article reports results on embedding sparse graph structures with characteristic properties into feed-forward networks and gives first insights into our study on network properties of sparse ANNs.
\textbf{Our Contributions} comprise
\begin{itemize}
	\item a technique to embed Directed Acyclic Graphs into Artificial Neural Networks (ANN)
	\item a comparison of Random Graph Generators as generators for structures of ANNs
	\item a performance estimator for ANNs based on structural properties of the networks' underlying graph
\end{itemize}

\subsection*{Related Work}
A lot of related works in fields such as network science, graph theory, and neuroevolution give \textbf{motivation} to study Sparse Structured Neural Networks (SNNs).
From a network science perspective, SNNs are Directed Acyclic Graphs (DAGs) in case of non-recurrent and Directed Graphs in case of Recurrent Neural Networks.
Directed acyclic and cyclic graphs are built, studied and characterized in graph theory and network science.
Notably, already Erd{\"o}s ``aimed to show [..] that the evolution of a random graph shows very clear-cut features'' \cite{erds1960evolution}.
Graphs with distinct degree distributions have been characterised as \textit{small-world} and \textit{scale-free} networks with the works of Watts \& Strogatz \cite{watts1998collective} and Barab{\'a}si \& Albert \cite{albert2002statistical}.
Both phenomens are ``not merely a curiosity of social networks nor an artefact of an idealized model -- it is probably generic for many large, sparse networks found in nature'' \cite{watts1998collective}.
However, from a biological perspective, the question of how the human brain is organised, still remains open:
According to Hilgetag et al. ``a reasonable guess is that the large-scale neuronal networks of the brain are arranged as globally sparse hierarchical modular networks'' \cite{hilgetag2016brain}.

Concerning the combination of \textbf{ANNs and Network Science} Mocanu et al. claim that ``ANNs perform perfectly well with sparsely-connected layers'' and introduce a training procedure \textit{SET}, which induces sparsity by magnitude-based pruning and randomly adding connections within the training phase \cite{mocanu2017evolutionary}.
While their method is driven by the same motivation and inspiration, it is conceptually fundamentally different from our idea of finding characteristic properties of SNNs.

The work of Bourely et al. can be considered very close to our idea of creating SNNs before the training phase.
They ``propose Sparse Neural Network architectures that are based on random or structured bipartite graph topologies'' \cite{bourely2017sparse} but do not seem to base their construction on existing work in Network Science when transforming their randomly generated structures into ANNs.

There exists a vast amount of articles on automatic methods which yield sparsity.
Besides well-established \textbf{regularisation} methods (e.g. $L_1$-regularisation), new structural regularisation methods can be found:
Srinivas et al. ``introduce additional gate variables to perform parameter selection'' \cite{srinivas2017training} and Louizos et al. ``propose a practical method for $L_0$ norm regularisation'' in which they ``prune the network during training by encouraging weights to become exactly zero'' \cite{louizos2017learning}.
Besides regularisation one can also achieve SNNs through pruning, construction and evolutionary strategies.
All of those domains achieve successes to some extent but seem to fail in providing explanations for why certain found architectures succeed and others do not.

The idea of \textbf{predicting the model performance} can already be found in a similar way in the works of Klein et al., Domhan et al. and Baker et al..
Klein et al. exploit ``information in automatic hyperparameter optimization by means of a probabilistic model of learning curves across hyperparameter settings'' \cite{klein2016learning}.
They compare the idea with a human expert assessing the course of the learning curve of a model.
Domhan et al. also ``mimic the early termination of bad runs using a probabilistic model that extrapolates the performance from the first part of a learning curve'' \cite{domhan2015speeding}.
Both are concerned with predicting the performance based on learning curve, not on structural network properties.

Baker et al. are closest to our method by ``predicting the final performance of partially trained model configurations using features based on network architectures, hyperparameters, and time-series validation performance data'' \cite{baker2018accelerating}.
They report good $R^2$ values (e.g. 0.969 for Cifar10 with MetaQNN CNNs) for predicting the performance and did so on slightly more complex datasets such as Cifar10, TinyImageNet and Penn Treebank.
However, our motivation is to focus only on architecture parameters and include more characteristic properties of the network graph than only ``including total number of weights and number of layers'' which is independent of other hyperparameters and saves conducting an expensive training phase.

~

\noindent In the following chapters we give an introduction to Network Science, illustrate how Directed Acyclic Graphs are embedded into ANNs, provide details on generated graphs and their properties and visualize results from predicting the performance of built ANNs only with features based on structural properties of the underlying graph.

\section{A Network Science Perspective: Random Graph Generators}
A graph $G = (V, E)$ is defined by its naturally ordered vertices $V$ and edges $E \subset V\times V$.
The number of edges containing a vertex $v\in V$ determines the degree of $v$.
The ``distribution function $P(k)$ [..] gives the probability that a randomly selected node has exactly $k$ edges'' \cite{albert2002statistical}.
It can be used as a first characteristic to differ between certain types of graphs.
For a ``random graph [it] is a Poisson distribution with a peak at $P(\langle k\rangle)$'', with $\langle k\rangle$ being ``the average degree of the network'' \cite{albert2002statistical}.
Graphs with degree distributions following a power-law tail $P(k) \sim k^{-\gamma}$ are called \textit{scale-free graphs}.
Graphs with ``relatively small characteristic path lengths'' are called \textit{small-world graphs} \cite{watts1998collective}.

Graphs can be generated by Random Graph Generators (RGG).
Various RGGs can yield very distinct statistical properties.
The most prominent model to generate a random graph is the \textit{Erdős-Rényi- / Gilbert}-model (ERG-model) which connects a given number of vertices randomly. 
While this ERG-model yields a Poisson distribution for the degree distribution, other models have been developed to close the gap of generating large graphs with distinct characterstics found in nature.
Notably, the first RGGs producing such graphs have been the \textit{Watts-Strogatz}-model \cite{watts1998collective} and the \textit{Barabási-Albert}-model \cite{albert2002statistical}.

\section{Embedding Arbitrary Structures into Sparse Neural Networks}
After obtaining a graph with desired properties by a Random Graph Generator, this graph is embedded into a Sparse Neural Network with following steps:
\begin{enumerate}
	\item Make graph directed (if not directed, yet),
	\item compute a layer indexing for all vertices,
	\item embed layered vertices between an input and output layer meeting the requirements of the dataset.
\end{enumerate}

\noindent The first step conducts a \textbf{transformation into a Directed Acyclic Graph} following an approach in Barak et al. \cite{barak1984maximal} which ``consider the class $\mathcal{A}$ of random acyclic directed graphs which are obtained from random graphs by directing all the edges from higher to lower indexed vertices''.
Given a graph and a natural ordering of its vertices, a DAG is obtained by directing all edges from higher to lower ordered vertices.
This can be easily computed by setting the upper (or lower) triangle of the adjacency matrix to zero.

\noindent Next, a \textbf{layer indexing} is computed to assign vertices into different layers within a feed-forward network.
Vertices in a common layer can be represented in a unified layer vector.
The indexing function $ind^{l}(v): V \rightarrow \mathbb{N}$ is recursively defined by $v \mapsto max(\{ind^{l}(s) ~|~ (s,v) \in E^{in}_v\} \cup \{-1\}) + 1$ which then defines a set of layers $\mathcal{L}$ and a family of neurons indexed by $I_l$ for each layer $l\in\mathcal{L}$.

\noindent Finally, the \textbf{layered DAG can be embed into an ANN} for a given task, e.g. a classification task such as MNIST with 784 input and 10 output neurons.
All neurons of the input layer are connected to neurons representing the vertices of the DAG with in-degree equal to zero (layer index zero).
Succedingly, each vertex gets represented as a neuron and is connected according to the DAG.
The build process finishes by connecting all neurons of the last layer of the DAG with neurons of the output layer.
Following this approach, each resulting ANN has at least two fully connected layers in size depending on the DAG vertices of in-degree and out-degree equalling zero.

\begin{figure}[tb]
	\centering
	\subfloat[Sketched graph from Random Graph Generator (RGG).]{\includegraphics[width=.3\textwidth]{\imagePath 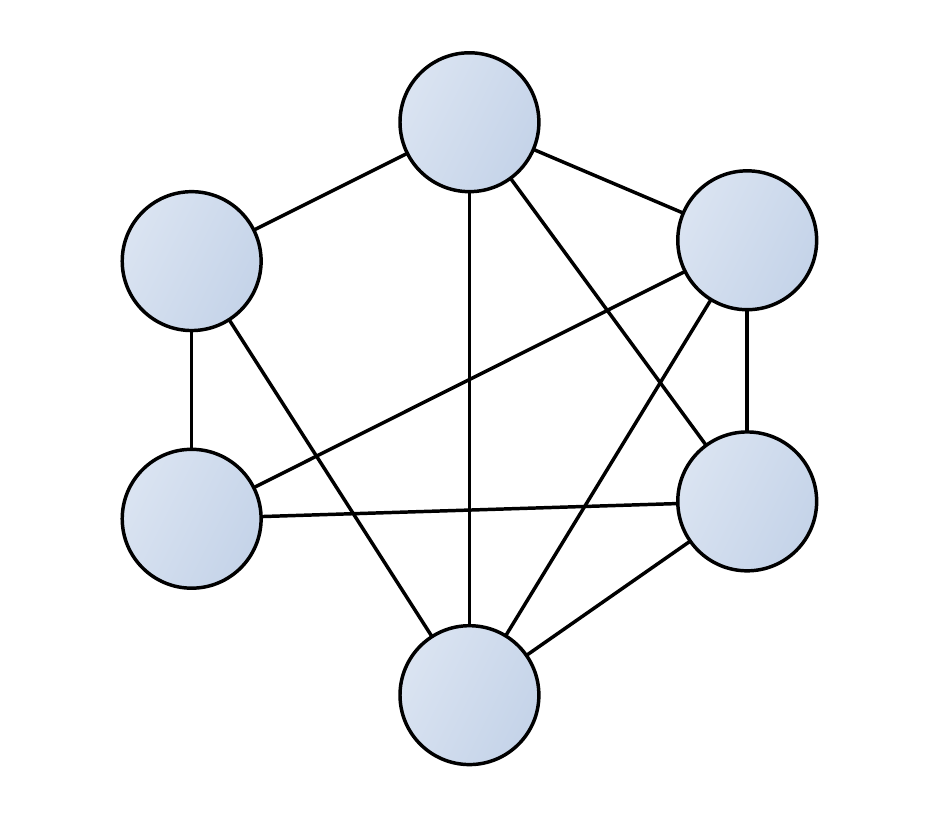}}
	\hfill
	\subfloat[Transformed Directed Acyclic Graph (DAG) from RGG.]{\includegraphics[width=.3\textwidth]{\imagePath 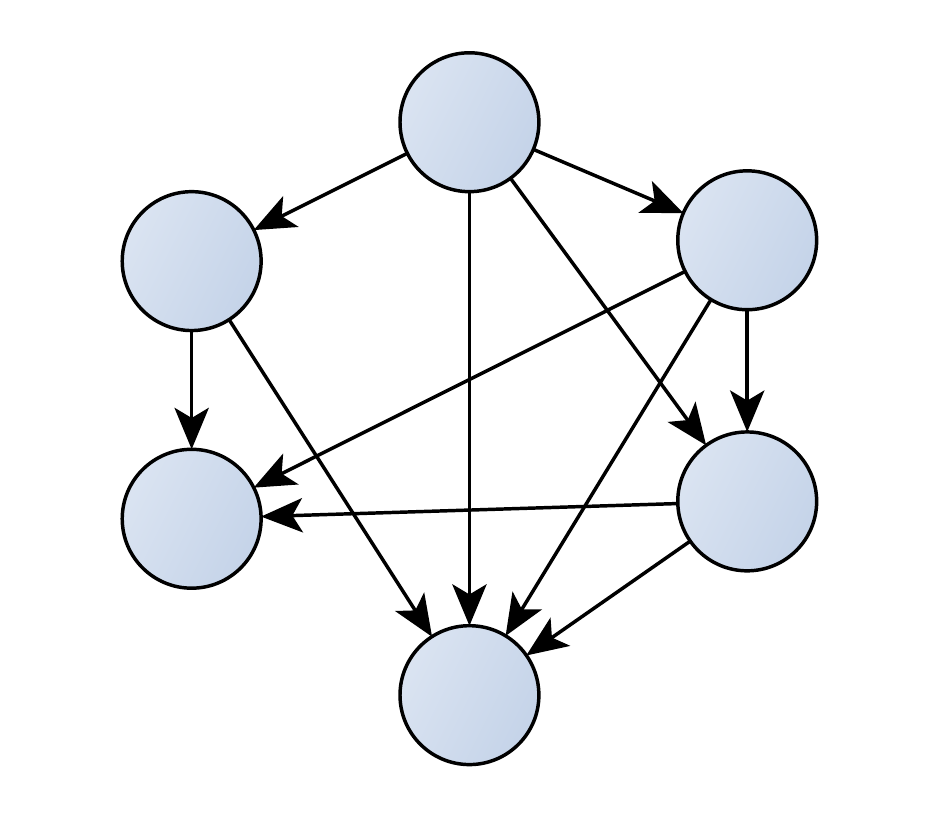}}
	\hfill
	\subfloat[Embedded DAG between input and output layers of an Artificial Neural Network classifier.]{\includegraphics[width=.3\textwidth]{\imagePath 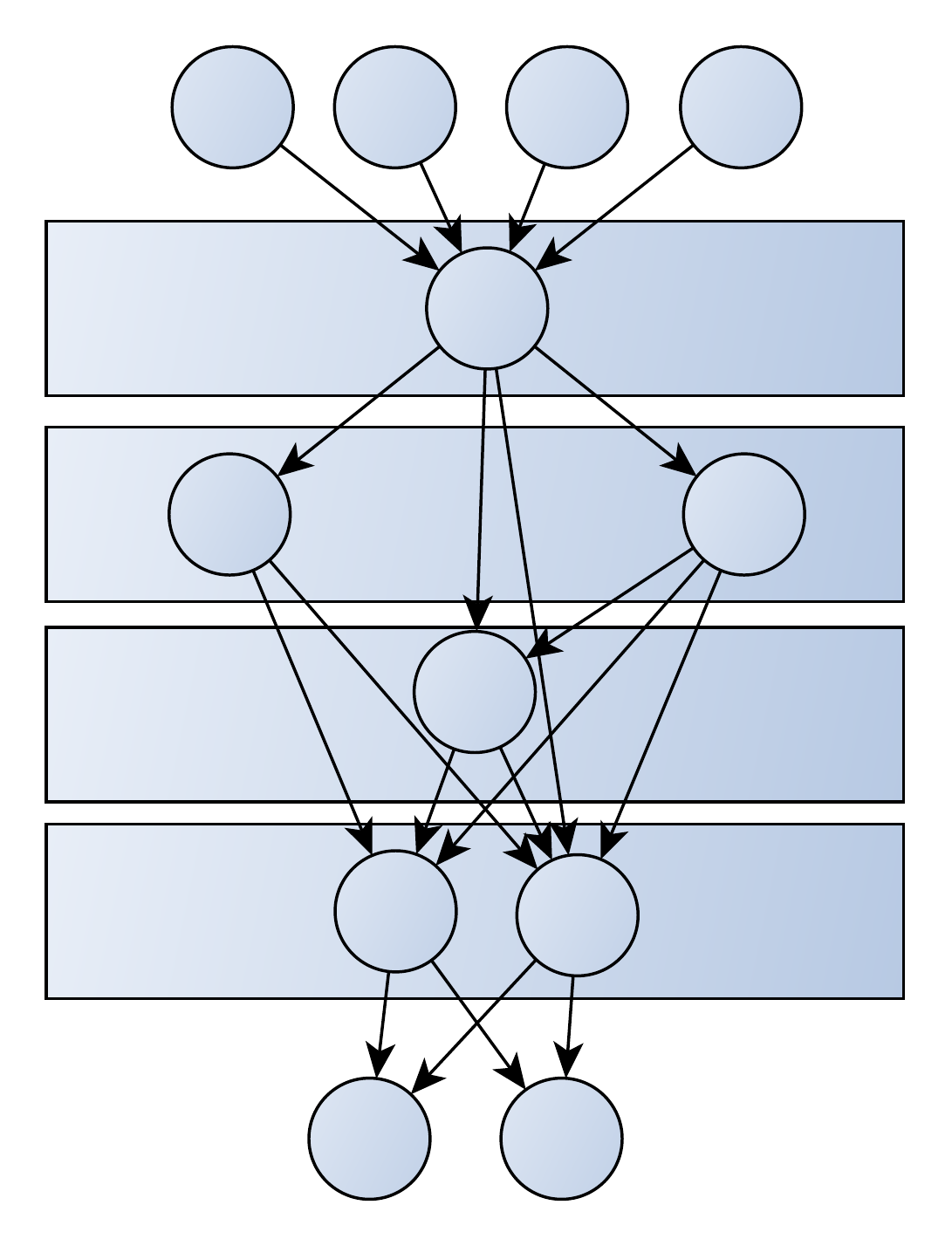}}
	\caption{
		Random (possibly undirected) graph, a directed transformation of it and the final embedding of the sparse structure within a Neural Network Classifier.
		In this example the classifier would be used for a problem with four input and two output neurons.
	}
	\label{fig:structural-integration}
\end{figure}

\autoref{fig:structural-integration} visualizes the steps from a random generated graph to a DAG and embedded into an ANN classifier.

\section{Performance Estimation Through Structural Properties}
In order to analyse the impact of the different structural elements, we conducted a supervised experiment predicting network performances on the structural properties only.
We then analysed the most important features involved in the decision along with the prediction quality.

For this experiment we created an artificial dataset \texttt{graphs10k} based on two Random Graph Generators (RGG).
Each graph in the dataset is transformed into a Sparse Neural Network (SNN), implemented in PyTorch \cite{paszke2017automatic}.
The resulting model is then trained and evaluated on MNIST \cite{lecun1998mnist}.
Based on obtained evaluation measures, three estimator models -- Ordinary Linear Regression, Support Vector Machine and Random Forest -- are trained by splitting \texttt{graphs10k} into a training and test set.
The estimator models give opportunity to discuss influence of structural properties to the SNN performances.

\subsection{The \texttt{graphs10k} dataset}
\begin{figure}[tb]
	\subfloat[Frequency distribution of number of edges.\label{fig:graphs10k-frequency-edges}]
	{\includegraphics[width=.48\textwidth]{\imagePath 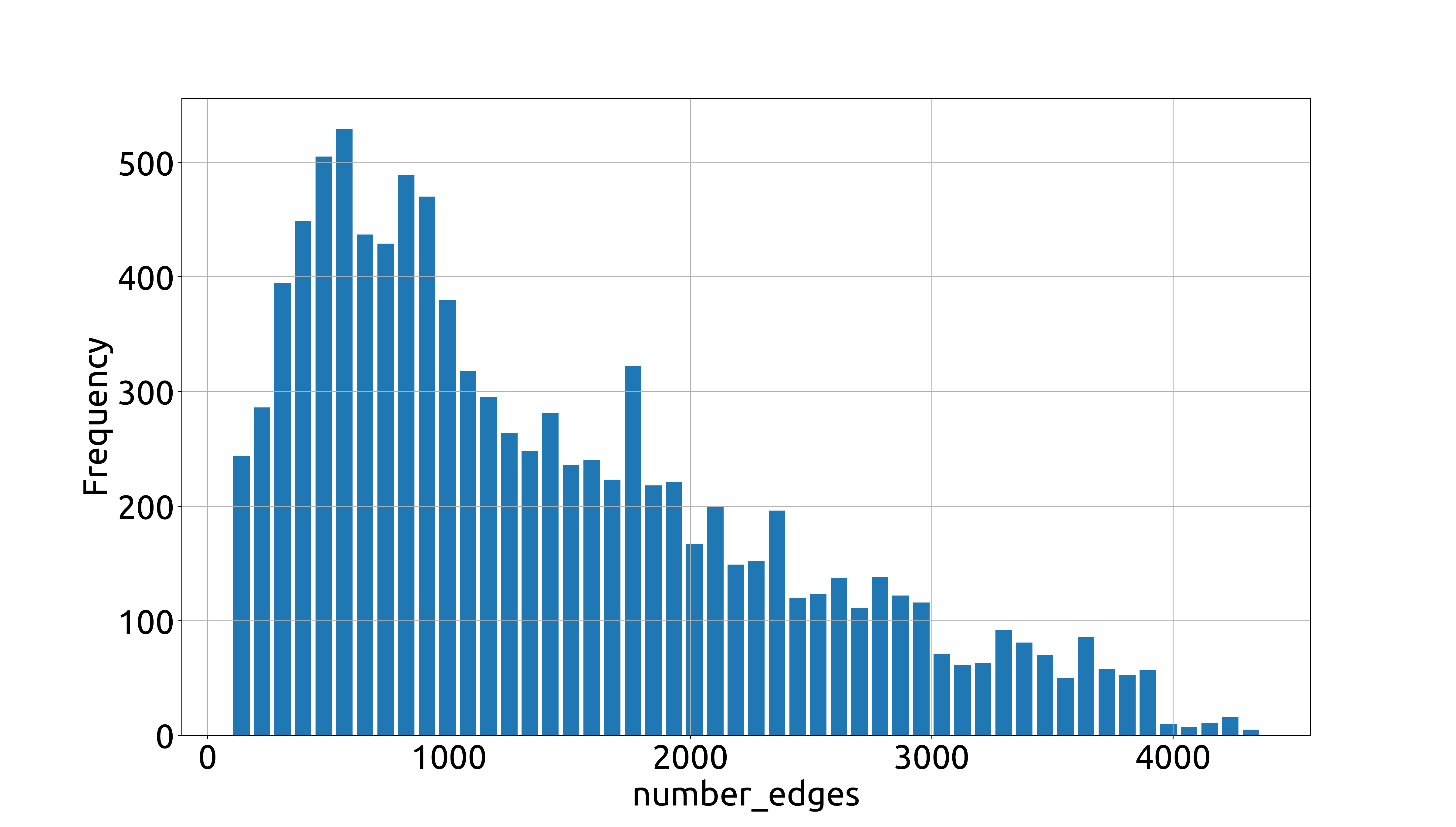}}
	\hfill
	\subfloat[Jointplot of eccentricity variance vs. test accuracy\label{fig:graphs10k-jointplot-eccentricityvar}]
	{\includegraphics[width=.48\textwidth]{\imagePath 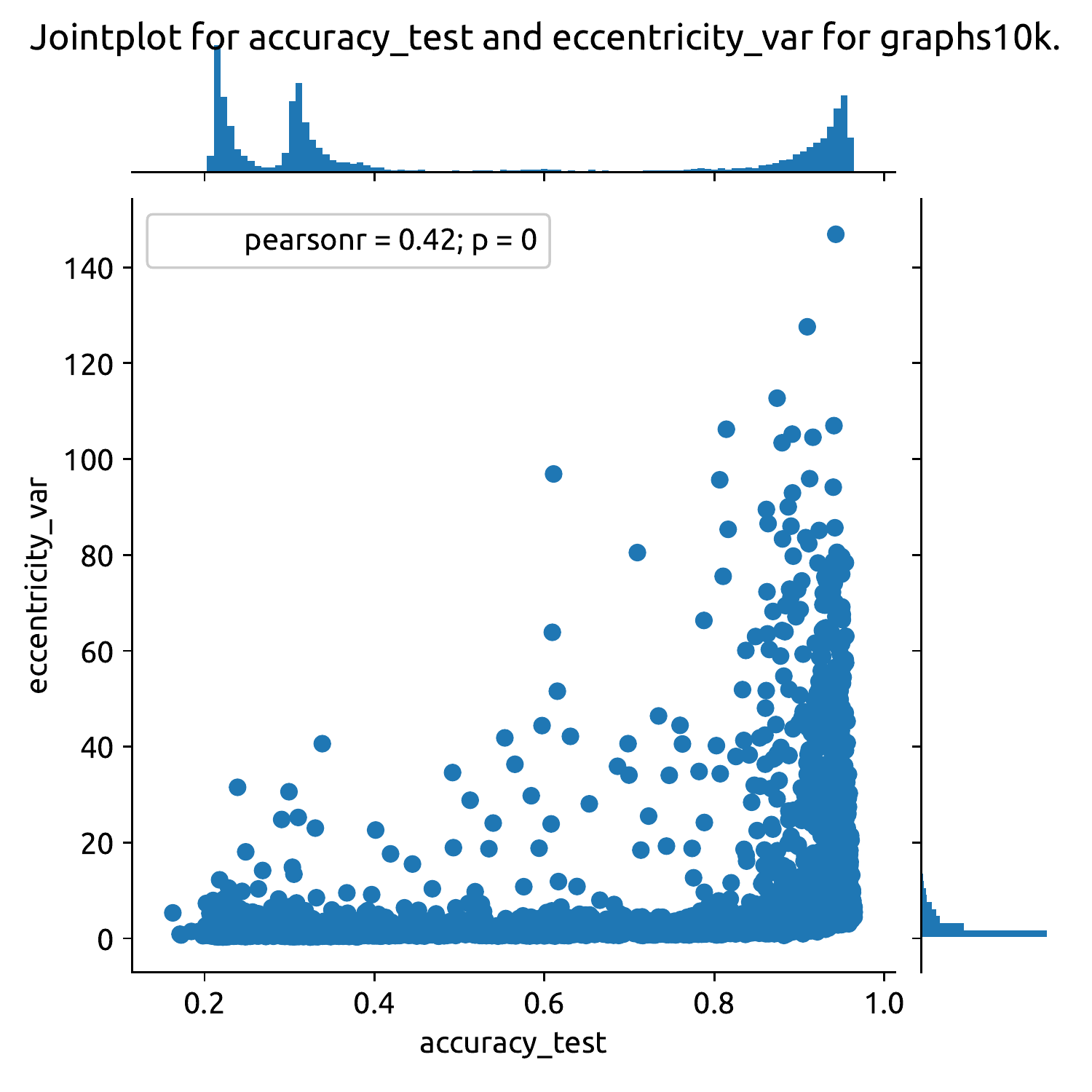}}
	\caption{\textbf{\autoref{fig:graphs10k-frequency-edges}}: An exemplary frequency distribution of \texttt{graphs10k}.
		With a uniformly distributed number of vertices, very different frequency distributions occur for other structural properties.
		\textbf{\autoref{fig:graphs10k-jointplot-eccentricityvar}}: Correlation between variance in eccentricity and accuracy shown in a joint plot.
		Axes contain distributions of each feature.
		The test accuracy distribution (on the top y-axis) shows that there only exist few graphs between $0.4$ and $0.85$ but three peaks at around $0.2$, $0.35$ and $0.95$.
		For high accuracies an increasing variance in eccentricity can be observed.
		A low eccentricity variance, however, does not explain good performance on its own.}
\end{figure}

To investigate which structural properties influence the performance of an Artificial Neural Network most, we created a dataset of graphs generated by \textit{Watts-Strogatz}- and \textit{Barabási-Albert}-models.
The dataset comprises 10,000 graphs, randomly generated by having between 50 and 500 vertices.
An exemplary property frequency distribution of the number of edges is shown in \autoref{fig:graphs10k-frequency-edges}.
The non-uniform distribution visualizes the difficulty of uniformly sampling in graph space\footnote{Note, that according to Karrer et al. ``we do not at present know of any way to sample uniformly from the unordered ensemble'' in the context of samling from possibile orderings of random acyclic graphs \cite{karrer2009random2}.

``Bayesian networks are composed of directed acyclic graphs, and it is very hard to represent the space of such graphs. Consequently, it is not easy to guarantee that a given method actually produces a uniform distribution in that space.'' \cite{ide2002random}}.

The dataset contains 5018 Barabási-Albert-graphs and 4982 Watts-Strogatz-graphs.
The graphs have between 97 and 4,365 edges and on average 1,399.17 edges with a standard deviation of 954,12.
In estimating the model performances the number of source and sink vertices are very prominent.
Source vertices are those with no incoming edges, sink vertices those with no outgoing edges.
On average the graphs have 79.75 source vertices with a standard deviation of 80.69 and 9.56 sink vertices with a standard deviation of 17.53.

Distributions within the graphs are reduced to four properties, namely minimum, arithmetic mean, maximum and standard deviation or variance.
For the mean degree distribution, a vertex has on average 10.36 connected edges and this arithmetic mean has a standard deviation of 4.47.
The variance of the degree distributions is on average 174.46 with a standard deviation of 259.08.

After each graph was embedded in a SNN, its accuracy value for test and validation set was obtained and added to the dataset.
More detailed statistics on other properties are given in \autoref{tbl:property-statistics}.

With the presented embedding technique it could be found, that graphs based on the Barabási-Albert model could not achieve comparable performances to graphs based on the Watts-Strogatz model.
Only 2618 of Barabási-Albert models achieved over 0.3 in accuracy (less than 50\%).
Excluding the Barabási-Albert model did not change the estimation results in \autoref{sec:estimators}, therefore statistics are reported for all graphs combined.
In future, the dataset will be enhanced to comprise more RGGs.

The test accuracy has three major peeks at arond 0.15, 0.3 and 0.94 as can be seen in the frequency distribution on the x-axis of \autoref{fig:graphs10k-jointplot-eccentricityvar}, depicting a joint plot of the test accuracy and the variance of a graphs' eccentricity\footnote{The eccentricity of a vertex in a connected graph is the maximum distance to any other vertex.} distribution.

\subsection{Estimators on \texttt{graphs10k}}\label{sec:estimators}
Ordinary Linear Regression (OLS), Support Vector Machine (SVM) and Random Forest (RF) are used to estimate model performance values given their underlying structural properties.

The dataset was split into a train and test set with a train set size ratio of 0.7.
In case of considering the whole dataset of 10,000 graphs the estimator models are trained on 7000 data points.
The train-test-split was repeated 20 times with different constants used as initialization seed for the random state.

Different feature sets are considered to assess the influence of single features.

The set $\Omega$ denotes all possible features as listed in \autoref{tbl:structural-properties}, $\Omega_{np}$ contains all features except for the number of vertices, number of edges, number of source vertices and the number of sink vertices.
These four features directly indicate numbers of trainable parameters in the model (features with \textbf{n}o direct indication to the number of \textbf{p}arameters).
$\Omega_{op}$ contains \textbf{o}nly those four \textbf{p}roperties.

Features with variance information only are considered in $\Omega_{var}$, namely the variances of the degree, eccentricity, neighborhood, path length, closeness and edge betweenness distributions.
Compared to other properties of the distributions -- such as the average, minimum or maximum -- features with variance information have shown in experiments to have some influence on the network performance.

A manually selected set $\Omega_{small}$ contains a reduced set of selected features, namely \textit{number\_source\_vertices}, \textit{number\_sink\_vertices}, \textit{degree\_distribution\_var}, \textit{density}, \textit{neighborhood\_var}, \textit{path\_length\_var}, \textit{closeness\_std}, \textit{edge\_betweenness\_std}, \textit{eccentricity\_var}.
\autoref{tbl:structural-properties} provides an overview of used feature sets and resulting feature importances for the RF estimator.

OLS achieved a $R^2$ score of 0.8631 on average for feature set $\Omega$ over 20 repetitions with a standard deviation of 0.0042.
Pearson's r for OLS is on average $\rho =$ 0.9291.
SVM with RBF kernel achieved on average 0.9356 with 0.0018 in standard deviation.
RF achieved a $R^2$ score of 0.9714 on average with a standard deviation of 0.0009.
None of the other considered estimators reaches the RF-estimator predicting the model performances.

The feature importances of the RF estimator are calculated with \textit{sklearn} \cite{sklearn2011} where ``each feature importance is computed as the (normalized) total reduction of the criterion brought by that feature'' \footnote{Taken from \url{http://scikit-learn.org/stable/modules/generated/sklearn.tree.DecisionTreeRegressor.html} which references Breiman, Friedman, \textit{Classification and regression trees}, 1984}.
They are listed in \autoref{tbl:structural-properties} for each used feature in scope of the used feature set.
The number of sink vertices clearly have most influence to the performance of a MNIST-classifier.

\begin{landscape}
\begin{table}[htb]
	\centering

\begin{tabular}{|l|l|l|l|l|l|l|}
	\hline
	\textbf{Model} & \textbf{$\Omega$} & \textbf{$\Omega_{np}$} & \textbf{$\Omega_{op}$} & \textbf{$\Omega_{var}$} & \textbf{$\Omega_{small}$} & \textbf{$\Omega_{min}$} \\
	\hline
	\textbf{Random Forest $R^2$} & \textbf{0.9714} $\plusminus$0.0009 & 0.9314 $\plusminus$0.0031 & 0.9664 $\plusminus$0.0010 & 0.9283 $\plusminus$0.0032 & \textbf{0.9710 $\plusminus$0.0011} & \textbf{0.9706 $\plusminus$0.0009} \\
	OLS $R^2$ & 0.8631 $\plusminus$0.0042 & 0.8476 $\plusminus$0.0053 & 0.6522 $\plusminus$0.0089 & 0.5968 $\plusminus$0.0103 & 0.7211 $\plusminus$0.0072 & 0.6907 $\plusminus$0.0077 \\
	$\text{SVM}_{lin}$  $R^2$ & 0.8620 $\plusminus$0.0045 & 0.8463 $\plusminus$0.0054 & 0.6432 $\plusminus$0.0114 & 0.5551 $\plusminus$0.0164 & 0.6943 $\plusminus$0.0101 & 0.6676 $\plusminus$0.0111 \\
	$\text{SVM}_{rbf}$ $R^2$ & 0.9356 $\plusminus$0.0018 & 0.9235 $\plusminus$0.0028 & 0.8561 $\plusminus$0.0041 & 0.7827 $\plusminus$0.0092 & 0.8781 $\plusminus$0.0045 & 0.8604 $\plusminus$0.0033 \\
	$\text{SVM}_{pol}$  $R^2$ & 0.9174 $\plusminus$0.0025 & 0.8998 $\plusminus$0.0032 & 0.5668 $\plusminus$0.0065 & 0.6839 $\plusminus$0.0117 & 0.8421 $\plusminus$0.0053 & 0.7543 $\plusminus$0.0083 \\
	\hline
	\hline
	\textbf{Property} & \multicolumn{6}{l|}{\textbf{RF Feature Importance}} \\
	\hline
	number\_vertices & 0.0009 & ~ & 0.0045 $\plusminus$0.0002 & ~ & ~ & ~ \\
	number\_edges & 0.0013 & ~ & 0.0071 $\plusminus$0.0002 & ~ & ~ &  ~ \\
	number\_source\_vertices & 0.0636 & ~ & 0.0720 $\plusminus$0.0018 & ~ & 0.0643 $\plusminus$0.0019 & 0.0659 $\plusminus$0.0019 \\
	number\_sink\_vertices & 0.9124 & ~ & 0.9164 $\plusminus$0.0019 & ~ & 0.9137 $\plusminus$0.0019 & 0.9139 $\plusminus$0.0020 \\
	degree\_distribution\_mean & 0.0004 & 0.0083 $\plusminus$0.0054 & ~ & ~ & ~ & ~ \\
	degree\_distribution\_var & 0.0009 & \textbf{0.4582} $\plusminus$0.0973 & ~ & 0.3685 $\plusminus$0.0866 & 0.0024 $\plusminus$0.0002 & 0.0069 $\plusminus$0.0002 \\
	diameter & 0.0003 & 0.0008 $\plusminus$0.0001 & ~ & ~ & ~ & ~ \\
	density & 0.0007 & 0.0030 $\plusminus$0.0007 & ~ & 0.0072 $\plusminus$0.0007 & 0.0023 $\plusminus$0.0002 & ~ \\
	eccentricity\_mean & 0.0015 & 0.0047 $\plusminus$0.0006 & ~ & ~ & ~ & ~ \\
	eccentricity\_var & 0.0022 & \textbf{0.3025} $\plusminus$0.1006 & ~ & 0.3401 $\plusminus$0.0994 & ~ & ~ \\
	eccentricity\_max & 0.0003 & 0.0009 $\plusminus$0.0002 & ~ & ~ & ~ & ~ \\
	neighborhood\_mean & 0.0004 & 0.0050 $\plusminus$0.0046 & ~ & ~ & ~ & ~ \\
	neighborhood\_var & 0.0011 & \textbf{0.1417} $\plusminus$0.0646 & ~ & 0.2434 $\plusminus$0.0883 & 0.0025 $\plusminus$0.0001 & ~ \\
	neighborhood\_min & 0.0005 & 0.0272 $\plusminus$0.0104 & ~ & ~ & ~ & ~ \\
	neighborhood\_max & 0.0017 & 0.0071 $\plusminus$0.0035 & ~ & ~ & ~ & ~ \\
	path\_length\_mean & 0.0011 & 0.0045 $\plusminus$0.0013 & ~ & ~ & ~ & ~ \\
	path\_length\_var & 0.0013 & 0.0052 $\plusminus$0.0009 & ~ & 0.0133 $\plusminus$0.0018 & 0.0034 $\plusminus$0.0001 & 0.0067 $\plusminus$0.0002 \\
	closeness\_min & 0.0014 & 0.0067 $\plusminus$0.0028 & ~ & ~ & ~ & ~ \\
	closeness\_mean & 0.0010 & 0.0030 $\plusminus$0.0003 & ~ & ~ & ~ & ~ \\
	closeness\_max & 0.0013 & 0.0034 $\plusminus$0.0003 & ~ & ~ & ~ & ~ \\
	closeness\_std & 0.0021 & 0.0064 $\plusminus$0.0012 & ~ & 0.0166 $\plusminus$0.0017 & 0.0039 $\plusminus$0.0002 & 0.0066 $\plusminus$0.0002 \\
	edge\_betweenness\_min & 0.0001 & 0.0008 $\plusminus$0.0003 & ~ & ~ & ~ & ~ \\
	edge\_betweenness\_mean & 0.0011 & 0.0041 $\plusminus$0.0005 & ~ & ~ & ~ & ~ \\
	edge\_betweenness\_max & 0.0015 & 0.0038 $\plusminus$0.0004 & ~ & ~ & ~ & ~ \\
	edge\_betweenness\_std & 0.0011 & 0.0027 $\plusminus$0.0003 & ~ & 0.0109 $\plusminus$0.0011 & 0.0036 $\plusminus$0.0001 & ~ \\
	\hline
\end{tabular}

	\caption{Structural properties for the SNNs' underlying graphs from the \texttt{graphs10k} dataset and their influence as features for the Random Forest model under different feature sets.}
	\label{tbl:structural-properties}
\end{table}
\end{landscape}

Features, which do not directly indicate the number of parameters, can be seen in the second column for $\Omega_{np}$ and the three most important ones are highlighted in boldface.
The three most important features are the variances of degree, eccentricity and neighborhood distributions of the graph.
Considering six variance features together with the number of source and sink vertices leads to a $R^2$ value for RF of $0.9710 \plusminus 0.0011$, only deviating by $0.0004$ from the best average $R^2$ value of $0.9714 \plusminus 0.0009$.
Variance features alone already achieve an average $R^2$ of $0.9283 \plusminus 0.0032$ (see $\Omega_{var}$).
This gives indication, that e.g. in the case of eccentricity a higher diversity leads to better performance.
Higher variances in those distributions imply different path lengths through the network which can be found in architectures such as Residual Networks.

\section{Conclusion \& Future Work}
This work presented motivations and approaches to study Artificial Neural Networks (ANNs) from a network science perspective.
Directed Acyclic Graphs with characteristic properties from network science are obtained by Random Graph Generators (RGGs) and embedded into ANNs.
A dataset of 10,000 graphs with \textit{Watts-Strogatz}- and \textit{Barabási-Albert}-models as RGGs was created to base experiments on.
ANN models are trained on the presented structural embedding technique and performance values such as the accuracy on a validation set were obtained.

With both, structural properties and resulting performance values, three estimator models, namely an Ordinary Linear Regression, a Support Vector Machine and a Random Forest (RF) model are built.
The Random Forest model is most successful in predicting ANN model performances based on the structural properties.
Influence of the structural properties (features of the RF) on predicting ANN model performances is measured.

Clearly, the most important feature is the number of vertices and edges, directly determining the number of trainable parameters.
There is, however, indication that the variance of distributions of properties such as the eccentricity, vertex degrees and path lengths of networks have influence for high performant models.
This insight goes along with successful models such as Residual Networks and Highway Networks, which introduce more variance and thus lead to a more ensemble-like behaviour.

More such characteristic graph properties will help \textbf{explain} differences of various architectures.
Understanding the interaction of graph properties could also be \textbf{exploitet} in the form of structural regularization -- designing architectures or searching for them (e.g. in evolutionary approaches) could be driven by network scientific knowledge.

In future work, insights into the influence of structural properties on model performance will be hardened by more complex \textbf{problem domains} which assumably have more potential of revealing stronger differences.
The approach with structural properties will be extended to \textbf{recurrent architectures} to reduce the gap between DAGs used in this work and directed cyclic networks, as they are found in nature.
The performance prediction will also be integrated into \textbf{neuroevolutionary} methods, most likely leading to a performance boost by early stopping and improved regularized search.

\bibliographystyle{splncs04}
\bibliography{\contentPath bibliography}

\newpage
\appendix

\begin{figure}[tb]
	\subfloat[Correlation between test accuracy and number of vertices of the underlying graph.]
	{\includegraphics[width=.48\textwidth]{\imagePath 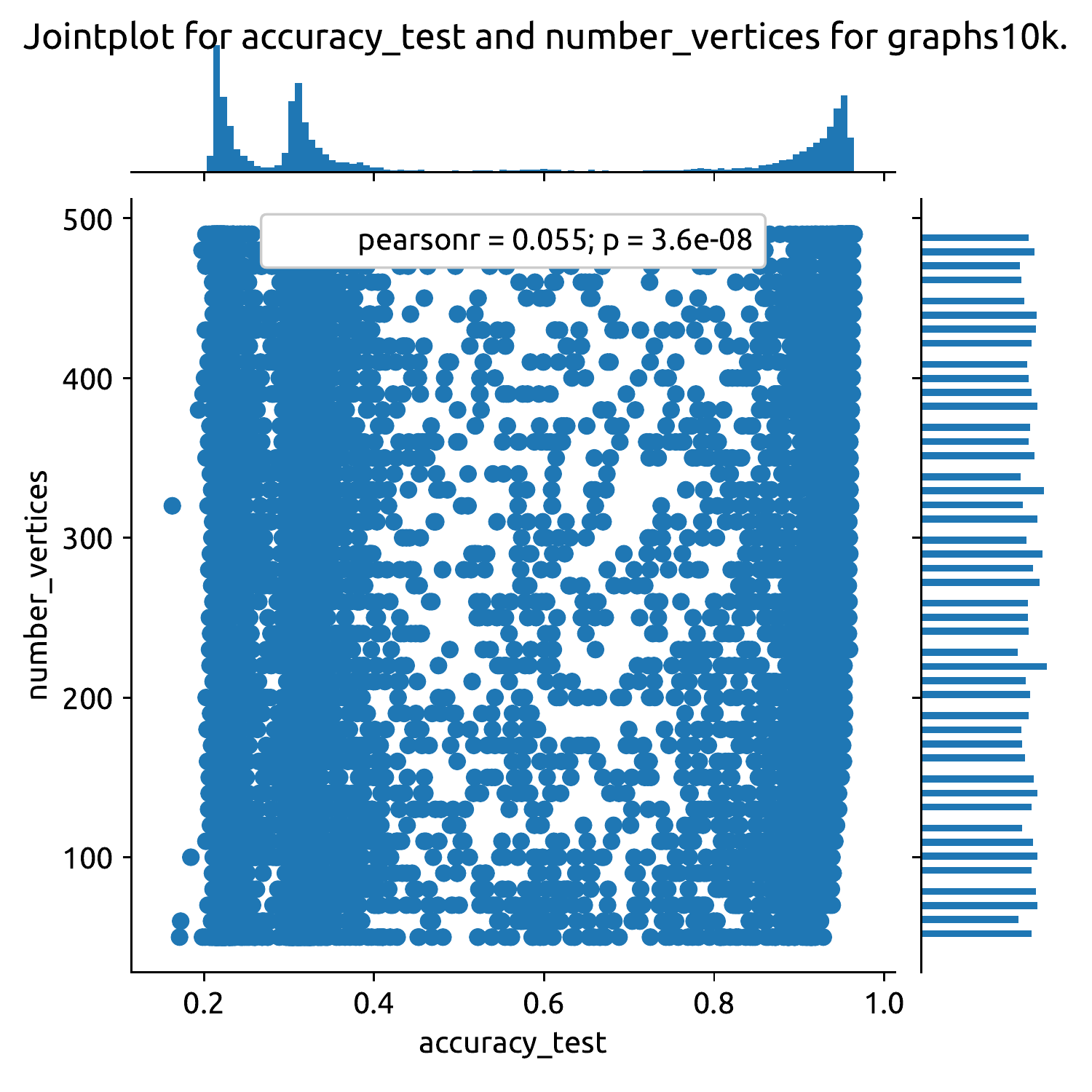}}
	\hfill
	\subfloat[Correlation between test accuracy and number of edges of the underlying graph.]
	{\includegraphics[width=.48\textwidth]{\imagePath 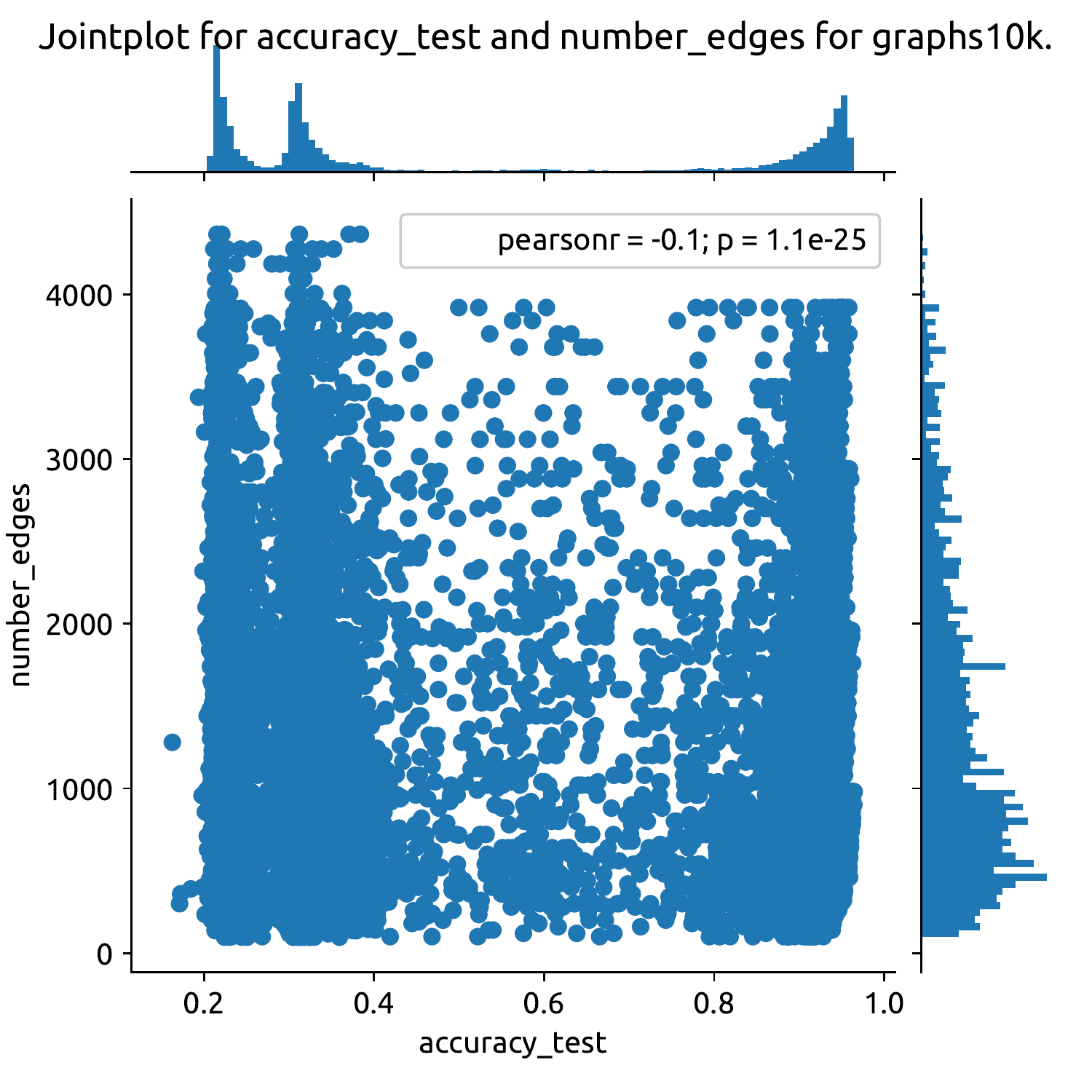}}
	\caption{}
\end{figure}

\begin{figure}[h!tb]
	\subfloat[Correlation between test accuracy and density of the underlying graph.]
	{\includegraphics[width=.48\textwidth]{\imagePath 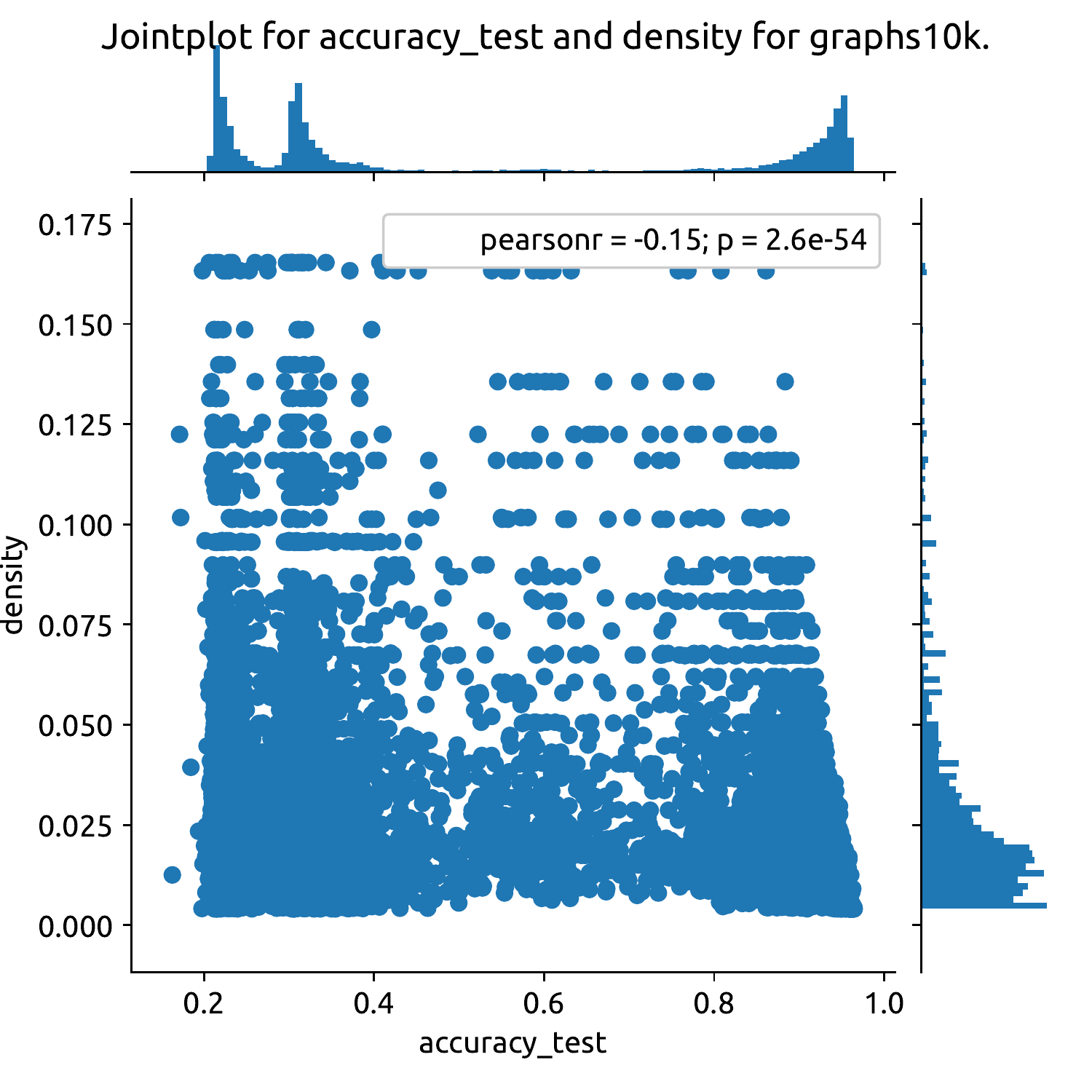}}
	\hfill
	\subfloat[Correlation between test accuracy and path length variance of the underlying graph.]
	{\includegraphics[width=.48\textwidth]{\imagePath 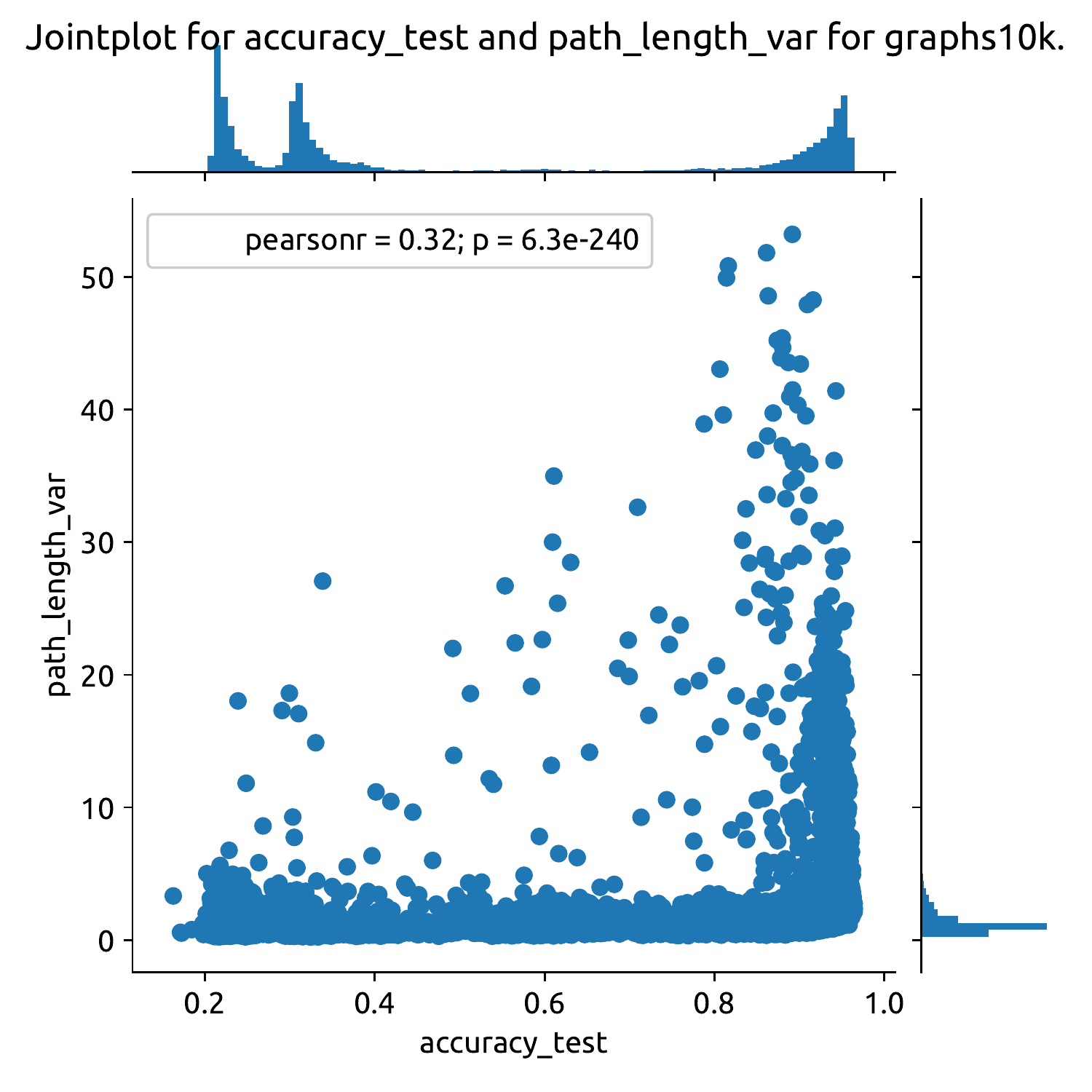}}
	\caption{}
\end{figure}

\begin{table}[htb]
	\centering
	\begin{tabular}{|l|l|l|l|l|}
		\hline
		\textbf{Property} & \textbf{Min} & \textbf{Mean} & \textbf{Max} & \textbf{Std} \\
		\hline
		number\_vertices & 50 & 269.66 & 490 & 129.56 \\
		number\_edges & 97 & 1399.17 & 4365 & 954.12 \\
		number\_source\_vertices & 1 & 79.7501 & 306 & 80.6862 \\
		number\_sink\_vertices & 1 & 9.5550 & 125 & 17.5298 \\
		diameter & 3 & 8.72 & 54 & 4.86 \\
		density & 0.0041 & 0.0277 & 0.1653 & 0.0256 \\
		\hline
		degree\_distribution\_mean & 3.88 & \textbf{10.36} & 17.82 & 4.47 \\
		degree\_distribution\_var & 0.4638 & 174.46 & 1274.97 & 259.08 \\
		\hline
		eccentricity\_mean & 1.7800 & \textbf{4.4104} & 29.2400 & 2.4913 \\
		eccentricity\_var & 0.2784 & 4.6160 & 146.8862 & 9.6251 \\
		eccentricity\_max & 3 & 8.7211 & 54 & 4.8585 \\
		\hline
		neighborhood\_mean & 4.8800 & \textbf{11.3557} & 18.8163 & 4.4670 \\
		neighborhood\_var & 0.4571 & 173.7798 & 1272.3148 & 258.3377 \\
		neighborhood\_min & 1 & 5.7161 & 14 & 2.8536 \\
		neighborhood\_max & 7 & 75.8607 & 312 & 72.7306 \\
		\hline
		path\_length\_mean & 1.3008 & \textbf{2.7941} & 14.2195 & 1.4005 \\
		path\_length\_var & 0.2229 & 1.9988 & 53.2144 & 3.8514 \\
		\hline
		closeness\_min & 0.0020 & 0.3229 & 0.5698 & 0.1275 \\
		closeness\_mean & 0.0476 & \textbf{0.4007} & 0.6150 & 0.1132 \\
		closeness\_max & 0.0514 & 0.5324 & 0.9672 & 0.1824 \\
		closeness\_std & 0.0090 & 0.0306 & 0.1137 & 0.0151 \\
		\hline
		edge\_betweenness\_min & 1 & 1.0034 & 3 & 0.0467 \\
		edge\_betweenness\_mean & 1.8272 & \textbf{45.2005} & 1531.7422 & 101.3437 \\
		edge\_betweenness\_max & 8.3333 & 441.7764 & 12259.0184 & 782.9232 \\
		edge\_betweenness\_std & 1.0408 & 51.3070 & 1781.8689 & 117.5626 \\
		\hline
	\end{tabular}
	\caption{Considered properties and their statistical distributions across the \texttt{graphs10k} dataset.}
	\label{tbl:property-statistics}
\end{table}

\end{document}